\documentclass[journel]{IEEEtran}
\usepackage{cite}
\usepackage{amsmath,amssymb,amsfonts}
\usepackage{algorithmic}
\usepackage{graphicx}
\usepackage{textcomp}
\usepackage{tikz}
\usepackage{caption}
\usepackage{multirow}
\usepackage{multicol}
\usepackage{epsf,color,url,booktabs,tabularx}

\begin{document}
	\title{SearchMorph: Multi-scale Correlation Iterative Network for Unsupervised Deformable Image Registration}
	\author{Xiao Fan, Shuxin Zhuang, Zhemin Zhuang, Ye Yuan, Shunmin Qiu, Alex Noel Joseph Raj, and Yibiao Rong
		\thanks{Xiao Fan is with the Shantou University, Shantou 515063, China (e-mail:20xfan1@stu.edu.cn). }
		\thanks{Shuxin Zhuang is with the Sun Yat-sen University, Shenzhen 518107, China (e-mail: shuxin613@qq.com).}
		\thanks{Ye Yuan and Alex Noel Joseph Raj are with the Shantou University, Shantou 515063, China (e-mail: sxzhuang@stu.edu.cn; yuanye@stu.edu.cn; jalexnoel@stu.edu.cn).}
		\thanks{Shunmin Qiu is with the First Affiliated Hospital of Shantou University,Shantou 515041, China (e-mail: shunmqiu@163.com).}
		\thanks{Yibiao Rong is with the Shantou University, Shantou 515063, China (e-mail:ybrong@stu.edu.cn).}
		\thanks{Corresponding author: Zhemin Zhuang is with the Shantou University, Shantou 515063, China (e-mail:zmzhuang@stu.edu.cn).}
		\thanks{This work was supported by the National Natural Science Foundation of China under Grant  82071992, Basic and Applied Basic Research Foundation of Guangdong Province under Grant  2020B1515120061, the Guangdong Province University Priority Field (Artifificial Intelligence) Project under Grant 2019KZDZX1013. }}
	\maketitle

\begin{abstract}
Deformable image registration can obtain dynamic information about images, which is of great significance in medical image analysis. The unsupervised deep learning registration method can quickly achieve high registration accuracy without labels. However, these methods generally suffer from uncorrelated features, poor ability to register large deformations and details, and unnatural deformation fields. To address the issues above, we propose an unsupervised multi-scale correlation iterative registration network (SearchMorph). In the proposed network, we introduce a correlation layer to strengthen the relevance between features and construct a correlation pyramid to provide multi-scale relevance information for the network. We also design a deformation field iterator, which improves the ability of the model to register details and large deformations through the search module and GRU while ensuring that the deformation field is realistic. We use single-temporal brain MR images and multi-temporal echocardiographic sequences to evaluate the model's ability to register large deformations and details. The experimental results demonstrate that the method in this paper achieves the highest registration accuracy and the lowest folding point ratio using a short elapsed time to state-of-the-art.
\end{abstract}


\section{Introduction}

Deformable image registration(DIR) is crucial in medical image processing and analysis. It maps a moving image onto a fixed image by looking for a spatial transformation. Traditional methods solve the image registration for a strategy of maximum or minimizing objective function\cite{ref33,ref34,ref35,ref36,ref37}.These methods often require much computation, and the registration process takes a long time. Moreover, traditional methods need to design different objective functions to re-fit different datasets, which leads to weak generalization ability of the model.

With the advent of deep learning, the application of deep learning in DIR has become a hot research topic. Image registration based on deep learning shows higher performance than traditional image registration methods and solves the problems of long registration time and weak generalization ability of traditional methods. Early supervised methods\cite{ref56,ref57,ref58,ref59} often used ground-truth deformation field as labels for image registration. However, supervised methods are challenging since the ground-truth deformation field is hard to obtain.

Unsupervised registration networks can carry out end-to-end learning without labels, overcoming the label dependence of supervised methods. However, these unsupervised methods also have some limitations. For example, most models based on the VoxelMorph framework directly predict the deformation field through feature maps, which makes the VoxelMorph framework-based models challenging to estimate large deformations. \cite{ref23}uses a strategy of multi-cascade iterations to learn the deformation field incrementally to improve the registration ability of the model for large deformations. This strategy is feasible. However, since the recursive cascade does not strengthen the correlation between features, the deformation field becomes unnatural while improving the registration performance. For multi-temporal image registration, \cite{ref26} proposes a joint learning framework for multi-temporal images, which achieves high scores on short-axis MR sequences by simultaneously optimizing segmentation and motion estimation branches. However, this method is challenging to register images with low signal-to-noise ratios, such as ultrasound images. Moreover, the joint learning framework cannot estimate the motion accurately without the segmentation branch.

In this work, we proposed an unsupervised multi-scale correlation iterative registration network (SearchMorph). The proposed model enables accurate registration of single-temporal MR images of the brain and accurate motion estimation of multi-temporal echocardiograms without adding any constraints.

The main contributions of our work are summarized as follows:

\begin{itemize}
	\item In this paper, a correlation layer is introduced, which calculates the cost volume of features, and the cost volume stores the correlation information between features. A multi-scale correlation pyramid is also constructed by pooling the last two dimensions of the cost volume. The correlation pyramid provides both small and large deformation information for the network, which enhances the ability of the network to register multi-scale deformation. The correlation pyramid also provides reference information for the network when registering images with low SNR.
	\item A deformation field iterator is designed, which uses GRU as the iterative module. GRU enables the network to register multiple times in a single prediction iteratively. The point to be registered gradually converge to a definite point by iteration. GRU refines the deformation field without using more parameters, helping the network learn helpful information in each iteration and improving the performance of model registration details and large deformable.
	\item  In this paper, a search module is also proposed for the deformation field iterator. The function of the search module is to search the correlation pyramid with a fixed radius to find the best registration point. This strategy ensures that the search range is from small to large. Based on this design, the model-registered image has better detail, and the deformation field is realistic.
\end{itemize}

\begin{figure*}
	\centering
	\includegraphics[scale=0.24]{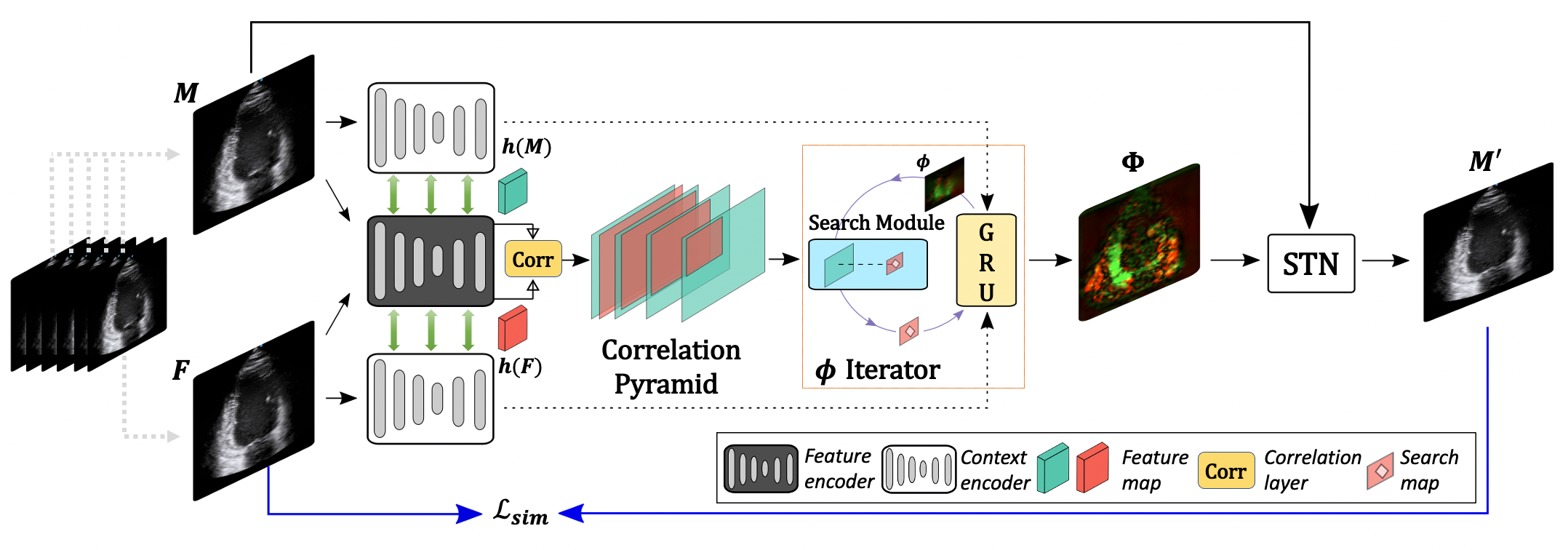}
	\caption{Overview of SearchMorph for medical image deformation registration. The network consists of four main components, (1) a feature encoder and two context encoders, which share weights; (2) a Correlation layer, which calculates the cost volume between features and constructs the correlation pyramid by pooling; (3) a Deformation field iterator, including GRU and search module, to optimize the deformation field by iteration; (4) STN, $\Phi$ warps $M$ by STN to obtain a warped image $M^{'}$ for back propagation, optimizing the whole network.}
	\label{Fig_1}
\end{figure*}

\section{Related work}
\subsection{Traditional image registration}

Classical medical registration methods usually use an iterative optimization strategy to minimize the objective function. These methods register two or more images by finding the spatial transformation between the images.

For deformable image registration, \cite{ref39,ref40,ref41} proposes elastic and B-splines models for the registration of multiple feature points of the image. The diffeomorphic transformation can solve the problem of overlap after the registration of pixel points. The Symmetric image normalization method (SyN) \cite{ref33} preserves the binary transformation of the topology. \cite{ref38} proposes the large displacement diffeomorphic metric mapping (LDDMM) to solve large displacement registration. \cite{ref37} proposes that the RDMM model tracks deformation through a special regularizer. Although these methods have made progress in image registration, each pair of images needs to establish a new objective function for registration mapping, which makes the registration process inefficient and time-consuming.

\subsection{Deep learning for image registration}

Image registration method based on deep learning has recently been a hot research topic. The registration time of deep learning methods is short, and due to their inductive nature, only one model can be used to register multiple images. Most people try supervised learning with segmentation labels or based on synthetic deformation fields as ground truth. However, the registration performance of these methods relies on the quality of the labels and synthetic deformation fields, which puts limitations on the information learned by the model.

The unsupervised method overcomes the reliance of the supervised methods on ground truth. Earliest, JaderBerg {\it et al.} proposed the spatial transformer network (STN) \cite{ref18}. STN automatically performs affine transformations on the input data without learning any parameters and has subsequently become an important part of the unsupervised framework. On this basis, Vos {\it et al.} uses the properties of the STN for unsupervised non-rigid body registration \cite{ref19}. VoxelMorph, proposed by Balakrishnan {\it et al.} enables unsupervised registration of brain MR data.  Xu {\it et al.} proposed a recursive cascade network VTN to improve the performance of unsupervised registration \cite{ref23}. VTN allows the model to progressively learn the deformable field by warping the image several times and thus registering it with the fixed image. Fan {\it et al.} used a discriminant instead of a loss function for registration \cite{ref24} . Zhu {\it et al.} used image pyramids for registration to register echocardiograms\cite{ref25}. Qin {\it et al.} proposed a joint learning network\cite{ref26} that registers time-series images by optimally segmenting branches and motion estimation. Kim {\it et al.} used cycle consistency to bring the images closer to diffeomorphic homogeneity\cite{ref43}. Chen {\it et al.} used the transformer encoder instead of the CNN\cite{ref104}.

\subsection{Optical Flow Estimation}

Optical flow estimation and image registration correlate the varying parts of two images. The difference is that optical flow often estimates the motion of a rigid object, whereas registration often estimates the deformation of a non-rigid object. Multi-temporal registration can be performed using the idea of motion estimation, which means that every pixel is tracked, thereby improving the performance of multi-temporal registration.

Dosovitskiy {\it et al.} proposed FlowNet\cite{ref9}, the first end-to-end deep learning method to predict optical flow fields. Then, Sun{\it et al.} proposed PWC-Net\cite{ref10}, which uses coarse-to-fine multi-scale feature combined with Cost Volumn\cite{ref11} for multiscale estimation of the optical flow field. Teed{\it et al.} proposed RAFT\cite{ref14}, which overcomes the limitation that the coarse-to-fine multi-scale method cannot accurately estimate small displacements. For unsupervised optical flow estimation, Yu {\it et al.} used STN for unsupervised optical flow \cite{ref20}. Meister releases UnFlow\cite {ref21} treats optical flow estimation as an image reconstruction problem. Luo {\it et al.} proposed UpFlow\cite {ref110}, which designed a novel pyramid structure to avoid image damage during downsampling.

\section{METHOD}

The proposed SearchMorph, as illustrated in Fig. 1. The network aims to refine the deformation field through an iterative approach to register the moving image with the fixed image accurately. Specifically, We input the moving image $M$ and the fixed image $F$ into the feature and context encoder and output the feature map and context. Then we input the feature maps $h(M)$ and $h(F)$ into the association layer and calculate the cost volume of the two feature maps. We then pool the last two dimensions of the cost volume to construct a multi-scale correlation pyramid. The deformation field iterator ($\phi$ Iterator) iterates over the refinement of the deformation field ($\phi$) by entering the correlation pyramid and the context. The deformation field $\phi$ is 2x upsampled to $\Phi$ at the last iteration to restore the original map resolution. When back-propagating, $M^{'}$ and $F$ perform a similarity loss calculation $\mathcal L_{sim}$ to optimize the weights of the whole network so that $M^{'}$ becomes increasingly similar to $F$. We will describe each part in detail below.

%
%

\subsection{Feature extactor}

The proposed feature extractor in this paper consists of a feature encoder and a context encoder. We use both encoders for feature extraction. The feature encoder extracts the feature map to calculate the correlation, and the context extractor extracts the context for feature information supplementation. The context addresses the problem of the network becoming one-sided due to the loss of information about the original features caused by calculating correlations.

The feature encoder and the context encoder are similar in structure to U-Net\cite{ref103}, with a skip-connect structure. The last layer of the feature encoder outputs an 8-channel feature map, while the context encoder outputs a 32-channel context. In the feature encoder, we split the feature map of the last layer into two 4-channel feature maps $h(F)$ and $h(M)$ corresponding to the input $F$ and $M$. Both encoders share weights for each layer except for the last layer. After each convolution, we normalize by Batch normalization and activate with Leakey relu. The exact structure of the feature extractor is shown in the appendix.

\subsection{Correlation Pyramid}

We introduce a correlation layer and construct a correlation pyramid. Correlation pyramids can store multi-scale correlation information and play two prominent roles in the network. Firstly, when registering images with complex deformation and low signal-to-noise ratio, the correlation pyramid provides the network with more information about the correlation of features. Secondly, in a subsequent step, the search module can combine multi-scale information for searching, improving the performance of model registration for multi-scale deformations.

The two feature maps are fed into the correlation layer, calculating the cost volume between the feature maps. The inner product of two feature maps can determine the feature correlation, often referred to as the cost volume calculation or affinity calculation. we assume that the two feature maps $h(M)\in\mathbb{R} ^{H\times W\times D} $,$h(F)\in\mathbb{R} ^{H\times W\times D} $, where $H,W$ is the length and width of the feature map and $D$ is the number of channels of the feature. The correlation between these two feature maps is calculated as follows:

\begin{subequations} 
	\begin{align}
		&C_{ijkl}= \sum_{d}h(M)_{i,j,d}\cdot h(F)_{k,l,d}\tag {1}\\
		&\mathrm{C}(h(M),h(F))\in \mathbb{R} ^{H\times W\times H\times W} \tag {2}
	\end{align}
\end{subequations}

In the equation above, where $C$ denotes the calculation of the correlation of one point. $\mathrm{C}(h(M),h(F))$ denotes $h(M),h(F)$ the cost volume of the two feature maps $h(M),h(F)$. $d$ denotes the channel for each pair of points. $i,j$ and $k,l$ denote the coordinates in the moving feature map $h(M)$ and the fixed feature map $h(F)$, respectively. Since $h(M)$ and $h(F)$ are obtained from the same feature encoder, $i,j$ and $k,l$ are in the same coordinate domain. Each pair of feature maps has a correlation volume of $\mathrm {C} \in \mathbb{R} ^{H\times W\times H\times W} $, where the first two dimensions and the last two dimensions correspond to the Moving image and the Fixed image, respectively.

In the construction of the correlation pyramid, we pool the last two dimensions of $\mathrm {C}(h(M),h(F))$ with convolution kernels 1,2,4,8 respectively to obtain four correlation matrices $\left \{ C^{0},C^{1},C^{2},C^{3}\right \} $. We refer to these four matrices as the correlation pyramid. The correlation pyramid holds multi-scale information about Fixed Image and serves as a search library for subsequent search modules. In addition, the correlation pyramid retains the high-resolution information from the Moving Image, allowing our model to predict rapidly deforming objects such as echocardiograms.

\begin{figure}
	\centering
	\includegraphics[scale=0.22]{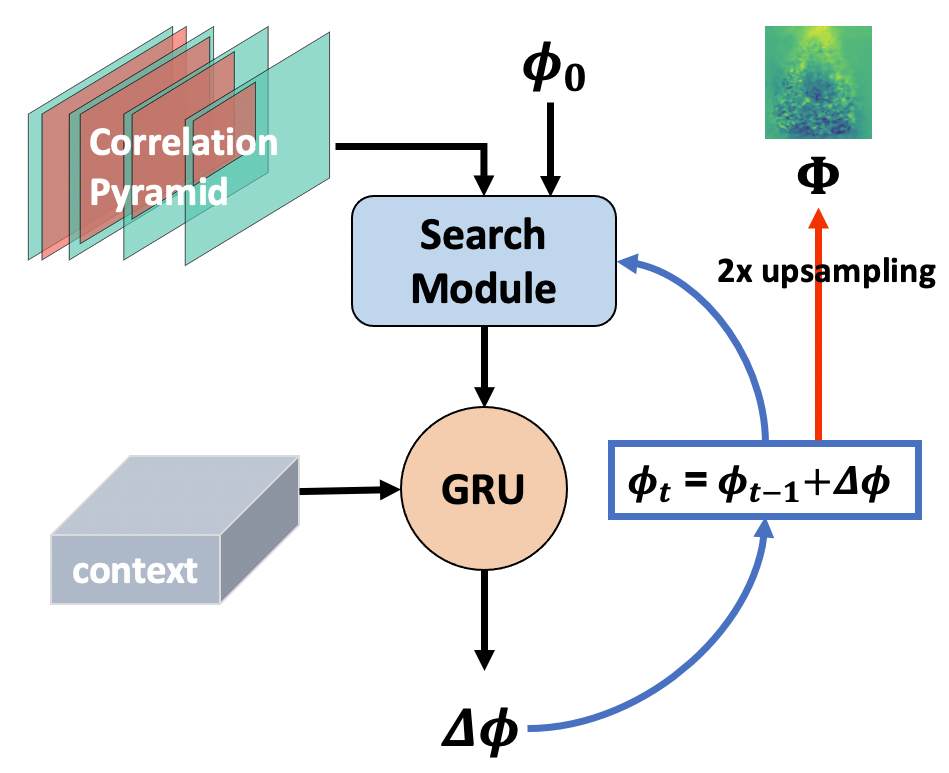}
	\caption{Figure of the iterative process of the deformation field iterator, $\phi_{0}$ denotes the input deformation field of the first iteration, $phi_{t}$ denotes the deformation field of this iteration, $\Delta \phi$ denotes the field displacement of this iteration, and $\Phi$ denotes the final output deformation field. SearchModule outputs the search map based on the correlation pyramid and $\phi_{t}$.  GRU outputs $\Delta \phi$ based on the search map and context information.}
	\label{Fig_2}
\end{figure}

\subsection{Deformation Field Iterator}

This paper proposes a deformation field iterator consisting of two parts, the search module, and the GRU. The search module performs a regional search of the correlation pyramid and outputs a search map. The search map stores information about the points around the point to be registered. GRU simulates the iterative process, which refines the deformation field through iterations. The search module allows for a clear strategy for model registration, alleviating the problem of unrealistic deformation fields. GRU iteratively refines the deformation field and enables the model to learn helpful information from each iteration. The deformation field iterator gives more detail to the aligned image and makes the deformation field dense and smooth.

The specific iteration steps of the deformation field iterator are shown in Figure 2. For the first iteration, we initialize the deformation field by $\phi_{0} = 0$ and then input the correlation pyramid and $\phi_{0}$ into the search module to output the search map. At each iteration, we input the search map and context into the GRU and output a deformation field update operator $\Delta \phi$. We can calculate the deformation field $\phi_{t}$ at the output of this iteration based on the update operator. As the iteration proceeds, the registration point gradually converges to a definite point. The deformation field is restored to the original map scale on the last iteration by 2x upsampling. 

\subsubsection{Search Module}

Suppose that the deformation field calculated in the last iteration is $\phi = (f^{x},f^{y})$. $f$ represents a matrix. The matrix holds the displacements of the deformation field, $f^{x}$ and $f^{y}$ hold the displacements in the $x$ direction and the displacements in the $y$ direction respectively. For a pixel $X = (u,v)$ in the first two dimensions of $\mathrm {C}(h(M),h(F))$, warping is performed using $\phi $. The warped pixel is $X^{w} = (u+f^{x}(u,v),v+f^{y}(u,v))$. We construct a set of neighbouring points $\mathcal P(X^{w})_{r}$ for $X^{w}$:

\begin{figure}
	\centering
	\includegraphics[scale=0.22]{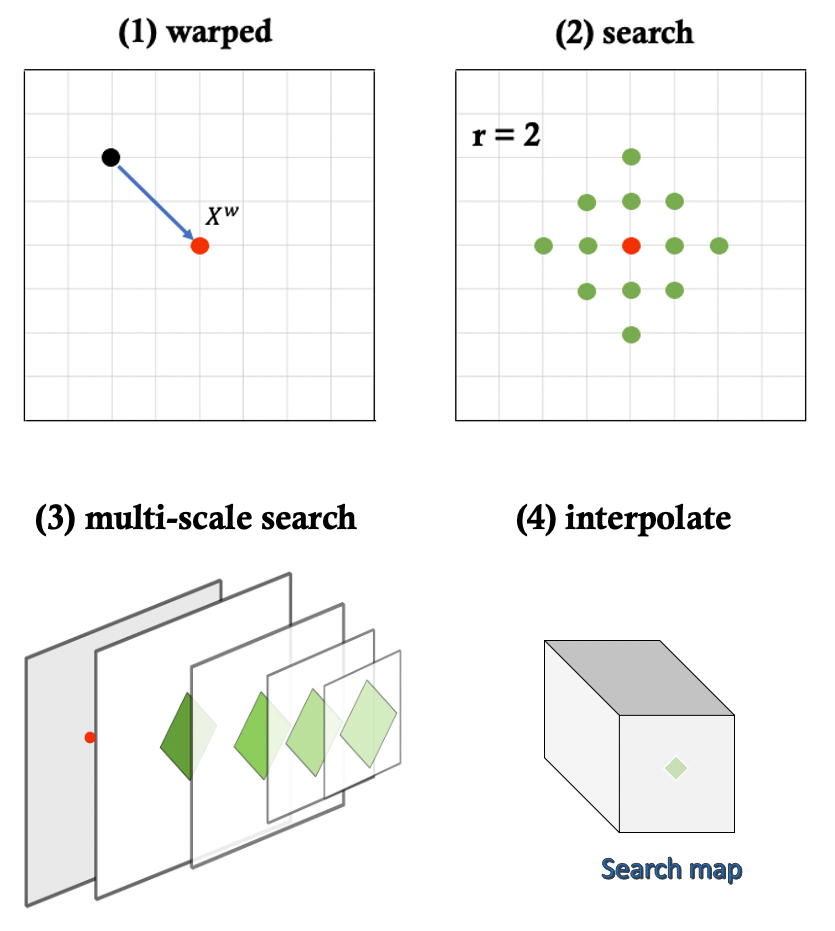}
	\caption{Process diagram of the search module. There are four steps in total, the first step uses the deformation field to warp the voxel, the second step performs a search range in the warped voxel points, the third step performs a search in the multiscale correlation pyramid, and the fourth step interpolates the multiscale map into a search map. We assume that the deformation field warps the black points to the red point $X^{w}$; the diamond-shaped region consisting of the green points is a search region with a search range of 2 pixels, and all green points are identified as having potential possible matches.}
	\label{Fig_2}
\end{figure}
\begin{equation}
	\mathcal P(X^{w})_{r} = \left \{  {X^{w} + D\mid D\in \mathbb{Z}^{2},\left \| D \right \|_{1}\le r}\right \} \tag {3}
\end{equation}

Take the L1 distance of radius $r$ as the search range of $X^{w} $ and define the neighborhood $\mathcal P(X^{w})_{r}$ of warped points according to this range. The correlation pyramid has four scales $\left \{ C^{0},C^{1},C^{2},C^{3} \right \} $. Map $X^{w} $ to each scale by interpolation, so each pixel point has a neighborhood of four scales. Searching across scales with a constant radius means a larger receptive field at lower scales. Finally, the values of each scale are concatenated into a feature map, called a search map or a motion map. The specific process of the search module is shown in Figure 3.

Theoretically, $r$ should be larger than the maximum deformation value between the two images. We believe that such a search strategy is similar to the diamond block matching method in traditional image processing \cite{ref52}. The difference is that block matching searches multiple times at the same scale, while this method searches once in each scale within one iteration.

\subsubsection{GRU}

The gated recurrent unit (GRU) is a recurrent neural network proposed to solve problems such as long-term memory. In the proposed model, we use the GRU to simulate the iterative recurrent step of a traditional alignment algorithm. GRU can select learning of helpful information in each iteration and allows the network to refine the deformation field multiple times in a single registration without using more parameters. The complete calculation process of GRU is as follows:

\begin{subequations} 
	\begin{align}
		r_{t}&=sigmoid(x_{t}W_{xr} + H_{t-1}W_{r} + b_{r})\tag {4}\\
		z_{t}&=sigmoid(x_{t}W_{xz} + H_{t-1}W_{z} + b_{z})\tag {5}\\
		\tilde{H}_{t}  &=tanh(x_{t}W_{hx} + R_{t}\odot H_{t-1}W_{h} + b_{h})\tag {6}\\
		H_{t}&=(1 - Z_{t}) \odot H_{t-1} + Z_{t} \odot \tilde{H}\tag {7}
	\end{align}
\end{subequations}
where $x_{t}$ denotes the input at the moment $t$, including the search map and context at the last moment. $h_{t}$ denotes the hidden state at moment $t$, $\tilde{h}_{t} $ denotes the hidden state of the candidate layer, and $Z_{t}$ denotes the update gate. The hidden state output by GRU is passed through two convolutional layers to predict the $\Delta \phi$ of the current update of the deformation field. Each iteration produces an update field, $\phi_{t}=\phi_{t-1}+\Delta \phi$. In the final output, the 2x upsampling deformation field restores it to the original map scale.

\subsection{Spatial transformer layer}

Spatial transformer network (STN)\cite{ref18} is a handy module, the full version of which can be placed in an arbitrary network to accomplish a certain degree of affine transformation, thus improving the predictive performance of the network.

In this paper, we introduce the latter two components of the STN, $Grid$ $generator$ and $Sampler$, to warp $M$. We call the Spatial transformer layer $\mathcal T$. After superimposing the deformed field, the original coordinate system is transformed into a warped image $M^{w} = T ( M,\phi)$  using a bilinear interpolation function. The equation for bilinear interpolation is:

\begin{equation}
	\mathcal T(M,\phi) = \sum_{q\in \mathcal N(p^{w})}M(q)\prod_{d\in \left \{  x,y\right \} }(1-\left |p^{w}_{d} -p_{d} \right | )\tag {8}
\end{equation}

Where $\mathcal N(p^{w})$ denotes the warped 4-coordinate neighborhood. $d$ denotes the two-dimensional space. The spatial transformer layer is invertible and does not have to learn any parameters, which can be trained end-to-end by back-propagation during the optimization process.

\subsection{Loss Function}

In deformable medical image registration, two steps are usually involved: a rigid transformation for global registration and a non-rigid transformation for local registration. The proposed network does not require a separate rigid transformation to obtain better results. The loss function of this network contains two components, the $\mathcal L_{sim} $ similarity loss term and the $\mathcal L_{reg} $ deformation field regularity term:

\begin{equation} 
	\mathcal L ( M ,  F ,\phi)=\mathcal L_{sim} (\mathcal T ( M,\phi),  F)+\alpha \mathcal L_{reg}(\phi)\tag {9}
\end{equation}

Where $ F$ denotes a fixed image. $ M$ denotes a moving image. $\phi$ denotes the deformation field of a pair of images. $\mathcal T$ represents the deformation, often referred to as $warp$\cite{ref51} in optical flow networks. In this network, the Spatial transformer layer takes on this part. In summary, $\mathcal L_{sim} $ measures how similar the deformed image $\mathcal T ( M,\phi)$ is to the similarity of the fixed image $ F$, $\mathcal L_{reg}$ penalizes the deformation field $\phi$ to make it smooth. $\alpha$ denotes the strength of the penalty term.

The similarity loss terms we use are mean square error $MSE$ and local normalized cross-correlation $LNCC$. Our experiments found that $MSE$ is more suitable for ultrasound modal images, and $LNCC$ would be more robust for more informative MR images.

The regular term is also known as the smoothing term. We use the most commonly used registration regular term, l2-loss, to penalize the deformation field. The penalized deformation field is smoothed so that the deformed image better matches the texture of the actual image.

\section{EXPERIMENTS}

\subsection{Datasets and Preprocessing}

We use four datasets to validate the validity of this method. They include a single-temporal brain MR dataset multi-temporal echocardiography dataset. The echocardiographic dataset consists of three different datasets. The details are described as follows:

LPBA40\cite{ref150} is a 3d brain MRI dataset. It contains brain MR images from 40 volunteers and is a single-temporal inter-patient dataset. As LPBA40 is 3D data, each of which has the format $160\times 192 \times 160 $ and contains 160 slices, we take the 80th slice of each case for registration in this paper.

CAMUS\cite{ref151} provides two-dimensional two- and four-chamber echocardiograms of five hundred patients, each with at least one entire cardiac cycle. This dataset provides manual labels for myocardial and cardiac blood pools at the end-diastolic ED and end-systolic ES, which we supplement with labels for the entire sequence.

Synthetic ground-truth data\cite{ref49} provides 105 sequence videos of A2C, A3C, and A4C with manual labels of the myocardial. We used the A3C in this dataset to compensate for the lack of three-chamber data and provide the cardiac blood pool as additional labels.

Echocardiography Video is the data we acquired. This dataset contains 20 patients, each containing at least two complete cardiac cycles, and we provide manual labels of the myocardial and cardiac blood pools for each frame.

\subsection{Metrics}

\begin{figure*}
	
	\centering
	\includegraphics[scale=0.08]{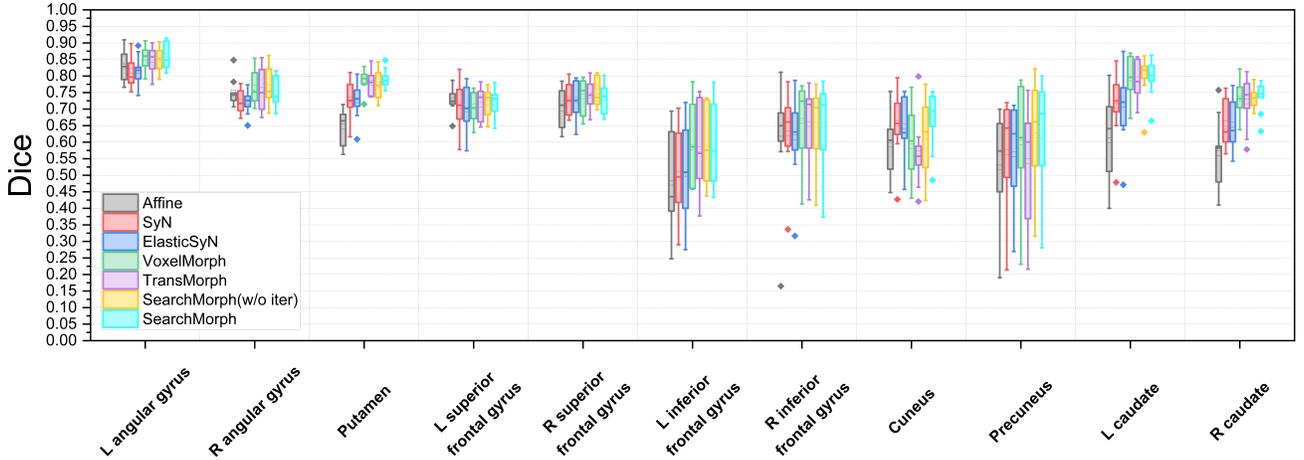}
	\caption{Brain multi-structure box line diagram. For quantitative evaluation of the registration ability of this paper's model and Baseline model for LPBA multi-structure. the Dice metric on the vertical axis and the multi-structure names on the horizontal axis, and the legend in the bottom left corner illustrating the method represented by each color.}
	\label{Fig_1}
\end{figure*}

We use three metrics to assess the registration ability of the model, the percentage of non-positive values in the determinant of the Jacobian matrix over the deformation field, the Dice index, and the registration time. We usually call the percentage of non-positive values in the determinant of the Jacobian matrix over the deformation field the folding point ratio, which does not make a distinction between the two later in the paper. RFP represents the ratio of the folding points of the deformation field to all voxel points, and it can measure the realism of the deformation field. The lower the folding point ratio, the closer the deformation field is to the diffeomorphism. The Dice index is used to determine the registration accuracy by comparing the overlap between the registered and manual labels; the lower the Dice index, the better the model performance. Time is the average time for each pair image registration.

\subsection{Implementation}
{
To compare model performance fairly, we used the learning rate of $1\times10^{-3}$ and the $Adam$ optimizer to update the weights in the neural network for each network. $\alpha$ in $MSE$ is set as 0.01 and $\alpha$ in $NCC$ is set as 2. During the training, the optimization step of each data is set as 1500 $epoch$, and $batch size$ is set as 8. We set the brain MR image size to $192 \times 160$and echocardiogram image size to $160 \times 160$. In the brain MR registration experiment, we set the search range $R$ of the model in this paper as 3. In the echocardiography experiment, we set $R$ as 2. We set the number of iterations of the deformation field iterator to 4

We have implemented all models on a Linux system with an Intel i7-11700K@3.60GHz x 16 processor and an NVIDIA GeForce RTX 3090 graphics card. The running environment is python version 3.7.11 and the PyTorch 1.8.0 framework, with Cuda version 11.1.1.
}

\subsection{Comparative experiments}

We use five state-of-the-art methods in the registration task as the baseline for our experiments to compare with our proposed model. The baseline includes three traditional methods, Affine, SyN by Advanced Normalization\cite{ref33} , ElasticSyN and two deep learning methods, VoxelMorph\cite{ref50}, Transmorph\cite{ref104}. In the experiments, SearchMorph(w/o iter) denotes the proposed model without iteration, and SearchMorph denotes the proposed model with four iterations. We validated the performance of the models using the brain MR dataset and the echocardiography dataset.

\begin{table}[]
	\centering
	\small
	\caption{The table shows the quantitative evaluation results of the LPBA brain MR slice registration. We use the average Dice index and the percentage of non-positive values in the determinant of the Jacobian matrix on the deformation field ($\%$ of $\left | J_{\phi} \right |\le 0$) for evaluation. The numbers in parentheses are the standard deviation between multiple data, reflecting registration stability. Bold indicates the highest score. SearchMorph(w/o iter) indicates the version of this model without iterations, and SearchMorph indicates the version of this model with four iterations.}
	\setlength{\tabcolsep}{1.2mm}{
		\begin{tabular}{l|ccc}
			\hline
			Model                                      & Dice                  & \multicolumn{1}{l}{$\%$ of  $\left | J_{\phi} \right |\le 0$} & \multicolumn{1}{l}{Time} \\ \hline
			Affine                                     & 0.639(0.036)          & -                     & 0.131                    \\
			SyN                                        & 0.675(0.039)          & \textless{}1e-4       & 0.316                    \\
			ElasticSyN                                 & 0.671(0.041)          & \textless{}1e-4       & 0.357                    \\ \hline
			VoxelMorph                                 & 0.709(0.030)          & 0.0027                & 0.012                    \\
			TransMorph                                 & 0.700(0.035)          & 0.0025                & 0.021                    \\ \hline
			\multicolumn{1}{c|}{SearchMorph(w/o iter)} & 0.715(0.029)          & 0.0005                & 0.027                    \\
			SearchMorph                                & \textbf{0.720(0.031)} & 0.0018                & 0.058                    \\ \hline
	\end{tabular}}
	
\end{table}
\begin{table*}[]
	\centering
	\small
	\caption{The table shows the results of the quantitative assessment of echocardiographic registration, and we used cardiac blood Dice and myocardial Dice as assessment metrics. The types of data evaluated include echocardiograms of 2CH, 3CH, and 4CH. Standard deviations are in parentheses. Bold indicates the highest score.}
	\setlength{\tabcolsep}{1mm}{
		\begin{tabular}{l|cc|cc|cc}
			\hline
			\multicolumn{1}{c|}{\multirow{2}{*}{Model}} & \multicolumn{2}{c|}{2CH}                                                   & \multicolumn{2}{c|}{3CH}                                                   & \multicolumn{2}{c}{4CH}                                                   \\ \cline{2-7} 
			\multicolumn{1}{c|}{}                       & \multicolumn{1}{l}{Blood pool Dice} & \multicolumn{1}{l|}{Myocardial Dice} & \multicolumn{1}{l}{Blood pool Dice} & \multicolumn{1}{l|}{Myocardial Dice} & \multicolumn{1}{l}{Blood pool Dice} & \multicolumn{1}{l}{Myocardial Dice} \\ \hline
			Affine                                      & 0.823(0.107)                        & 0.755(0.121)                         & 0.851(0.048)                        & 0.762(0.146)                         & 0.846(0.089)                        & 0.736(0.140)                        \\
			SyN                                         & 0.876(0.094)                        & 0.797(0.110)                         & 0.900(0.038)                        & 0.800(0.125)                         & 0.876(0.080)                        & 0.781(0.125)                        \\
			ElasticSyN                                  & 0.875(0.096)                        & 0.796(0.111)                         & 0.901(0.035)                        & 0.802(0.120)                         & 0.877(0.080)                        & 0.781(0.125)                        \\ \hline
			VoxelMorph                                  & 0.879(0.112)                        & 0.871(0.146)                         & 0.914(0.022)                        & 0.906(0.030)                         & 0.905(0.075)                        & 0.876(0.120)                        \\
			TransMorph                                  & 0.881(0.114)                        & 0.873(0.148)                         & 0.913(0.085)                        & 0.880(0.127)                         & 0.910(0.080)                        & 0.881(0.129)                        \\ \hline
			SearchMorph                                & \textbf{0.888(0.112)}               & \textbf{0.880(0.142)}                & \textbf{0.921(0.021)}               & \textbf{0.914(0.028)}                & \textbf{0.919(0.068)}               & \textbf{0.891(0.113)}               \\ \hline
	\end{tabular}}
\end{table*}
\subsubsection{Single-temporal Brain MR Registration}	
{Table I shows the quantitative results between the proposed method SearchMorph and the comparison method. As seen from the table, SearchMorph and SearchMorph(w/o iter) achieved the highest Dice score of 0.727 and the next highest of 0.723, respectively. SearchMorph and SearchMorph(w/o iter) also achieved the next lowest RFP of 0.018 and the lowest 0.0005, respectively. Even though the folding point rate increased after adding iterations, the overall level was still lower than other deep learning models. Even though the folding point rate increases after adding iterations, the overall level is still lower than other deep learning models. In terms of registration time, SearchMorph(w/o iter) is close to TransMorph's average image registration time per pair but slightly higher than VoxelMorph.  SearchMorph costs a longer time than other deep learning models due to the addition of iteration but still has a significant advantage over traditional models. In addition, we demonstrated in subsequent ablation experiments that the method in this paper has the potential to accomplish more accurate registration using less time.
	
Figure 4 shows the box line diagram of the results of the multi-structural registration of brain MR, which contains the registration information of our model with the Baseline model for 11 critical structures such as Cuneus and Precumeus. As seen from the box line plot, SearchMorph scores higher than other models in several structures and has higher stability and upper limits. In particular, SearchMorph excels in the Caudate nucleus and Angular gyrus. Since different structures have different sizes, the box line diagrams' results fully demonstrate the model's excellent performance in registering both large and small deformations.

\begin{figure}
	
	\centering
	\includegraphics[scale=0.32]{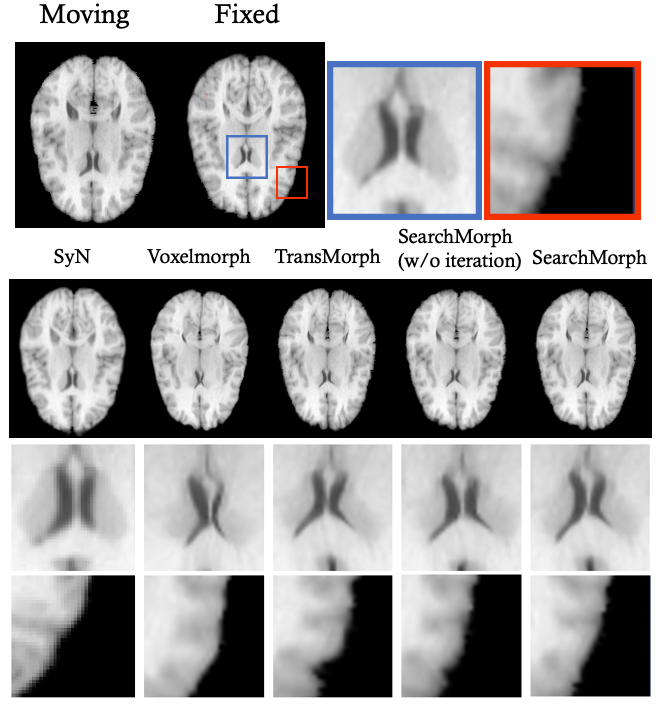}
	\caption{Example of LPBA brain MR slice registration. The blue box outlines the Lateral ventricle and Caudate nucleus, and the red box outlines the Subfrontal gyrus. The first row shows the moving image, fixed image, and fixed images in the blue and red boxes, the second row shows the results of image deformation for the four registration methods. The third and fourth rows show enlarged views of the four methods in the blue and red boxes, respectively.}
	\label{Fig_1}
\end{figure}

Figure 5 shows the comparison results of brain MR registration based on the proposed method and the baseline method. It can be seen from Fig. 5 that the SearchMorph-registered images are morphologically closer to the fixed images. To more clearly demonstrate the disparity between the methods in this paper and others, we have boxed the Lateral ventricle and Caudate nucleus in blue boxes and the Subfrontal gyrus in red boxes. The figure shows that the SearchMorph-registered image is closer to the Fixed image at both the blue and red boxes. The SyN-registered image has the correct shape in the blue box, but its warp size is incorrect. VoxelMorph and TransMorph are correct in size but not in shape. SearchMorph is correct in shape and size. The remaining three methods register incorrectly in the red box, and SearchMorph is similar to the Fixed image.}


\begin{figure*}
	
	\centering
	\includegraphics[scale=0.25]{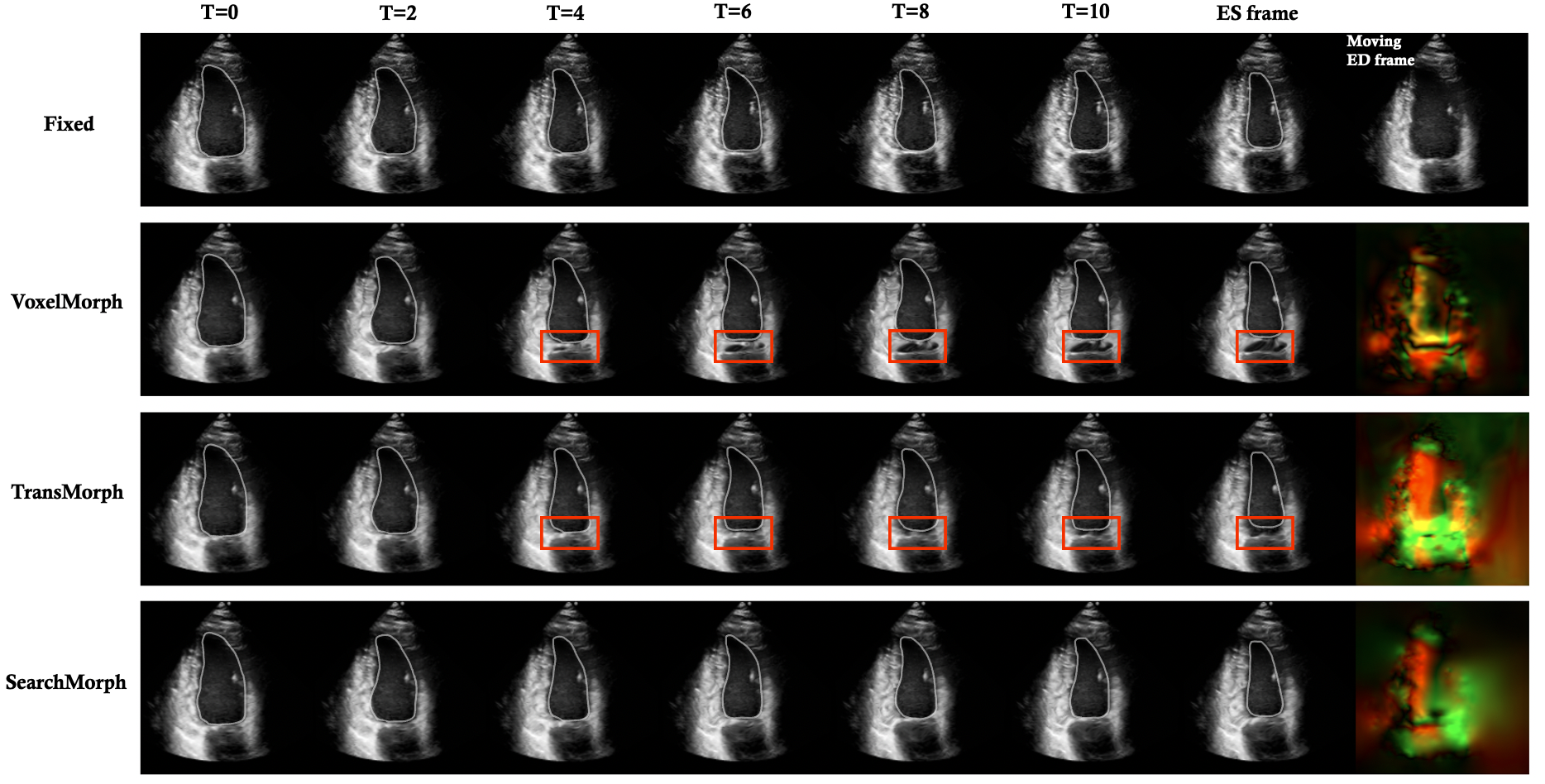}
	\caption{The effect of registration of each model in the systolic sequence of two-chamber echocardiography is shown. The fixed image t=n indicates the nth frame and T=12 is the ES frame, Moving image is the ED frame. Ground-truth with endocardium marked in gray. We frame in red the areas where the registration image deviates significantly from the Fixed image. The first row shows Fixed sequence images and the Moving image. The second, third, and fourth rows represent VoxelMorph, TransMorph, and SearchMorph registered echocardiograms, respectively, and their last column shows the deformation fields resulting from the registration of end-diastolic images with end-systolic images.}
	\label{Fig_1}
\end{figure*}

\subsubsection{Multi-temporal Echocardiogram registration}
{The echocardiogram registration experiments evaluate the model's ability to register multi-temporal and ultrasound modality data with many artifacts. The data used in this experiment include two-chamber, four-chamber CAMUS, three-chamber Synthetic, and our acquisition of two-chamber echocardiograms for a total of 1636 frames in 80 cases. We compare the registration performance of the Dice assessment model for myocardial and cardiac blood pools for this data.
	
Table II demonstrates the performance of proposed and baseline methods when registering the echocardiogram. We used the left ventricular cardiac blood pool DICE and myocardial DICE as evaluation metrics, and the size of the left ventricle varied in the echocardiography of the different chambers. The table shows that SearchMorph had the highest cardiac blood pool and myocardial Dice in two-chamber, three-chamber, and four-chamber echocardiograms. The experimental results demonstrate that the method in this paper can adapt to multi-scale deformation and shows excellent performance in the registration of echocardiography with multiple cavity hearts.
	
Figure 6 demonstrates the echocardiographic systolic registration. The Fixed image is a sequence of two-chamber echocardiograms, where T=12 is the image of the left ventricle at end-systole, and the Moving Image is the image of the left ventricle at end-diastole. We marked the ground truth of the endocardium as gray and compared it with the model-registered images to observe the registration effect. VoxelMorph and SearchMorph registrations are better during the first few frames, and TransMorph shows minor deviations. From the sixth frame onwards, the VoxelMorph and TranMorph registered images deviate significantly from the Fixed image, as shown in the red boxed area. VoxelMorph has a better registration of the inner membrane but is defective due to excessive registration deformation, and the endocardial of TransMorph-registered images do not fit the ground truth. The most significant deviations occur when registering the ES frame. Only the SearchMorph-registered echocardiographic sequence was highly overlapping with the endocardial ground truth. The experimental results demonstrate that our model performs best when registering echocardiograms, primarily when registering large deformations showing excellent performance.
	
The deformation field plots for the three methods are shown in the last column, red for moving to the right, green for moving to the left, and the shade of the color represents the distance moved. The echocardiogram shown is in systole, with its left wall moving to the right and its right wall moving to the left. As seen in the figure, the deformation field of VoxelMorph registration is almost haphazard, indicating that VoxelMorph does not have a clear registration strategy. The TransMorph-registered deformation field exhibits a direction of motion that generally conforms to the pattern of cardiac contraction, but its deformation field shows deformation spillover. The spillover manifests in a large red area on the left side that extends beyond the myocardial wall. The TransMorph-registered deformation field also has a red area on the right side, representing a significant deformation estimation error. SearchMorph produces a deformation field consistent with the contraction motion, and the deformation field is smoother and more accurately estimated.
}
\begin{table}[]
	\centering
	\small
	\caption{The table shows the results of the ablation experiments, where we evaluated the contribution of context, search module, correlation pyramid, GRU, and iteration to the model registration performance using brain MR data, and we used the mean Dice and the percentage of non-positive values in the determinant of the Jacobian matrix as an evaluation metric. Standard deviations are shown in parentheses.}
	\setlength{\tabcolsep}{1.5mm}{
		\begin{tabular}{l|ccl}
			\hline
			Model             & Dice                  & \multicolumn{1}{l}{$\%$ of  $\left | J_{\phi} \right |\le 0$}    & Time           \\ \hline
			w/o context       & 0.710(0.026)          & \textbf{\textless{}1e-4} & 0.058          \\
			w/o pyramid       & 0.714(0.027)          & 0.0013                   & 0.059          \\
			w/o search module & 0.704(0.040)          & 0.0020                  & 0.059          \\
			w/o GRU           & 0.708(0.032)          & 0.0015                   & \textbf{0.021} \\
			w/o iteration     & 0.715(0.029)          & 0.0005                   & 0.027          \\ \hline
			SearchMorph       & \textbf{0.720(0.031)} & 0.0018                   & 0.059          \\ \hline
	\end{tabular}}
\end{table}
%
\subsection{Ablation experiments}
{We designed a set of ablation experiments to validate the contribution and necessity of the SearchMorph vital components. These components include context encoder, search module, correlation pyramid, GRU, and iteration. We conducted experiments with brain MR data and judged the alignment performance of the model using the average Dice, the percentage of non-positive values in the determinant of the Jacobian matrix on the deformation field, and the average test time per image. 
	
Table III shows the results of the ablation experiments. We can see that the Dice of the model decreases when removing any component. Among them, the Dice drops the most when removing the search module and GRU module, which are 0.704 and 0.708, respectively, much smaller than the 0.720 of the full version of SearchMorph. This result shows that each network module designed in this paper has improved the registration accuracy. When extreme performance is not sought, the version removing iteration has a lower percentage of non-positive values in the determinant of the Jacobian matrix on the deformation field and shorter test times. After removing the context, correlation pyramid, GRU, and iteration, the folding point rate of the model also decreases. After removing the context, the percentage of non-positive values in the determinant of the Jacobian matrix is <1e-4 and exhibits diffeomorphic performance in this dataset. After removing the search module, the percentage of non-positive values in the determinant of the Jacobian matrix is 0.0020, which has increased. This result shows that the search module somewhat reduces the folding point rate. The ablation experiments prove the necessity of all the components proposed in this paper.}


\section{DISCUSSION}

\subsection{Comparative experiments}	
{SearchMorph achieved the highest average Dice and the lowest Jacobi determinant non-positive ratio in single-temporal brain MR experiments. SearchMorph not only shows excellent performance in registering multiple structures in the brain but also optimizes more registration details after adding iterations. We analyze this for three important reasons. (1) We designed the deformation field iterator to perform iterative registration in one prediction. The registration points converge gradually to a definite point. So the deformation field gradually approaches the actual value after adding iterations, optimizing more registration details. (2) We designed the correlation pyramid. The correlation pyramid enables the model to adapt to multiple displacement scales and provides more registration information to the network when registering a single image, improving the registration accuracy. (3) Our model achieves a lower RFP. This result is mainly due to the search strategy we defined for SearchMorph, which makes the model-registered deformation fields more realistic.
	
In multi-temporal echocardiographic experiments, SearchMorph performs best in two-chamber, three-chamber, and four-chamber hearts. The experimental results demonstrate the strong capability of the present model in registering multi-scale deformations. The SearchMorph-registered images best fit the myocardial endocardium of the Fixed images, demonstrating the ability of the model to register images with large deformations. We analyze this for four important reasons. (1) This model uses correlation pyramids as the search library for registration, which complements the multi-scale correlation information and enhances the ability of the network to register multi-scale deformations. (2) Due to the characteristics of low signal-to-noise ratio and many artifacts in ultrasound images, it is hard to accurately register the ultrasound images by conventional methods. Instead of using the conventional method of direct output of deformation fields through features, the method in this paper provides multi-scale correlation information for the network. This strategy can mitigate the interference of noise in feature extraction. (3) The search module designed in this paper searches the correlation pyramid in a fixed radius, ensuring that the module's search is small to large. This strategy ensures that the model can register accurately for large deformations at low resolutions. (4)  The SearchMorph registered image deformation field conforms to the direction of motion of the heart during systole, which is sufficient to demonstrate that the registration of two images evaluated by the similarity of grayscale values alone is not sufficient. In this paper, instead of using loss to restrict the model, we give the model a search strategy to make the deformation field more logical.
}

\subsection{Ablation experiments}	

{The results of the ablation experiments show that all structures of the present model contribute to the improved registration performance. We will analyze the reasons for this specifically. (1) In the ablation experiments of the context extractor, removing context information causes a decrease in the Dice score of the network. We believe that contextual feature information can complement each other with correlation information to improve the registration performance of the model, which is consistent with our original intention of including context. After removing the context, the percentage of non-positive values in the determinant of the Jacobian matrix on the deformation field <1e-4. One explanation is that context and correlation store information in different patterns, resulting in a slower network fit. (2) After removing the correlation pyramid, the Dice index decreases. We believe that the decreasing part is most likely the detailed part of the image. Because the correlation pyramid provides multi-scale correlation, it allows the network to synthesize more information for decision making, thus optimizing many registration details. At the same time, the percentage of non-positive values in the determinant of the Jacobian matrix on the deformation field decreases after removing the correlation pyramid. We believe the network will better register images when information increases, but it will also cause more learning stress. (3) With the search module removed, Dice drops to 0.704, a total of 0.016 less than the search module added. The search module does two main things, limiting the registration range and searching for registration points in a multi-scale pyramid, both of which are indispensable. Together, they make the network search carefully for a correct registration point within a range during the registration, improving the accuracy of the registration. With the removal of the search module, the percentage of non-positive values in the determinant of the Jacobian matrix on the deformation field increases. This result is because, in most datasets, the same part of the moving and fixed images are within a fixed range, and the search module gives a more precise target for the registration task. (4) In testing the contribution of GRU, we use three convolutional layers instead of GRU for iteration. The model's performance is worse when removing GRU, but it takes a short time. (5) SearchMorph shows only a 0.005 decrease in Dice without iteration, while the time doubles. As seen from the qualitative experiments, adding iteration optimizes more registration details, so iteration is necessary to achieve high performance.}

\subsection{Limitations}	

{The network proposed in this paper has some limitations. (1) Due to GPU memory limitations, the feature encoder in this paper only upsamples to 1/2 of the original image scale. A lower resolution will affect the registration to some extent. (2) The search strategy proposed in this paper is limited to two-dimensional data. In future work, we expect to design a three-dimensional search strategy adapted to three-dimensional images. (3) The proposed model, while maintaining a low folding point ratio, falls short of differential homozygosity. In subsequent work, we will design a diffeomorphic version of SearchMorph.}

\section{CONCLUSION}

There is a gradually growing consensus that it is difficult to register images with only one inference. Improving the registration performance through iteration has become a tough hot spot in registration research. We propose an unsupervised multi-scale correlation iterative registration network, SearchMorph. SearchMorph establishes links between features by calculating the cost volume between features and refining the deformation field in a deformation field iterator. We have also designed a search module that registers voxel points to their surrounding similarities, thus improving the accuracy of the registration. The experimental results demonstrate that the proposed model exhibits excellent performance in both single-temporal MR and multi-temporal ultrasound images and possesses a lower folding point ratio.

\end{document}